\documentclass[twocolumn, switch]{article} % Two-column formatting
\usepackage[table]{xcolor}		% colors

\usepackage{preprint}

\usepackage{amsmath, amsthm, amssymb, amsfonts}

\usepackage[backend=biber,
bibstyle=ieee,
citestyle=numeric-comp,
sorting=none,
isbn=false,
doi=true,
url=false,
eprint=true
]{biblatex}
\usepackage[utf8]{inputenc}	% allow utf-8 input
\usepackage[T1]{fontenc}	% use 8-bit T1 fonts

\usepackage{booktabs, tabularx, threeparttable, makecell, siunitx}% professional-quality tables
\usepackage{multirow}

\usepackage{nicefrac}		% compact symbols for 1/2, etc.
\usepackage{microtype}		% microtypography
\usepackage{lineno}		% Line numbers
\usepackage{float}			% Allows for figures within multicol
\usepackage{dblfloatfix}

\usepackage{lipsum}		%  Filler text

\usepackage{authblk}
\usepackage{abstract}
\usepackage[labelfont=bf]{caption}

\usepackage[colorlinks = true,
            linkcolor = purple,
            urlcolor  = blue,
            citecolor = cyan,
            anchorcolor = black]{hyperref}	% Color links to references, figures, etc.

\usepackage{newfloat}
\DeclareFloatingEnvironment[name={Supplementary Figure}]{suppfigure}
\usepackage{sidecap}
\sidecaptionvpos{figure}{c}

\usepackage{titlesec}
\titlespacing\section{0pt}{12pt plus 3pt minus 3pt}{1pt plus 1pt minus 1pt}
\titlespacing\subsection{0pt}{10pt plus 3pt minus 3pt}{1pt plus 1pt minus 1pt}
\titlespacing\subsubsection{0pt}{8pt plus 3pt minus 3pt}{1pt plus 1pt minus 1pt}

\definecolor{lightgray}{gray}{0.9}

\begin{document}

%%%%%%%%%%%%%%%%   Title   %%%%%%%%%%%%%%%%
\title{Forecasting the Progression of Alzheimer's Disease Using Neural
Networks and a Novel Pre-Processing Algorithm}

%%%%%%%%%%%%%%%  Author list  %%%%%%%%%%%%%%%

\renewcommand*{\Authfont}{\bfseries}
\author{Jack Albright\authorcr\normalfont for the Alzheimer's Disease Neuroimaging Initiative\thanks{Data used in preparation of this article were obtained from the Alzheimer’s Disease Neuroimaging Initiative (ADNI) database (adni.loni.usc.edu). As such, the investigators within the ADNI contributed to the design and implementation of ADNI and/or provided data but did not participate in analysis or writing of this report. A complete listing of ADNI investigators can be found at: \url{http://adni.loni.usc.edu/wp-content/uploads/how_to_apply/ADNI_Acknowledgement_List.pdf}}}
\affil{The Nueva School, San Mateo, CA}

%%%%%%%%%%%%%%    Front matter    %%%%%%%%%%%%%%

\twocolumn[ 
  \begin{@twocolumnfalse} 
  
\maketitle

\begin{abstract}
Alzheimer's disease (AD) is the most common neurodegenerative disease in
older people. Despite considerable efforts to find a cure for AD, there
is a 99.6\% failure rate of clinical trials for AD drugs, likely because
AD patients cannot easily be identified at early stages. This project
investigated machine learning approaches to predict the clinical state
of patients in future years to benefit AD research. Clinical data from
1737 patients was obtained from the Alzheimer's Disease Neuroimaging
Initiative (ADNI) database and was processed using the ``All-Pairs''
technique, a novel methodology created for this project involving the
comparison of all possible pairs of temporal data points for each
patient. This data was then used to train various machine learning
models. Models were evaluated using 7-fold cross-validation on the
training dataset and confirmed using data from a separate testing
dataset (110 patients). A neural network model was effective (mAUC =
0.866) at predicting the progression of AD on a month-by-month basis,
both in patients who were initially cognitively normal and in patients
suffering from mild cognitive impairment. Such a model could be used to
identify patients at early stages of AD and who are therefore good
candidates for clinical trials for AD therapeutics. 
\end{abstract}
%\keywords{First keyword \and Second keyword \and More} % (optional)
\vspace{0.35cm}

  \end{@twocolumnfalse} 
] 

\saythanks

%%%%%%%%%%%%%%%  Main text   %%%%%%%%%%%%%%%
% \linenumbers

\section{Introduction}
Alzheimer's disease (AD) is the most common neuro-degenerative disease
in older people. \cite{Selkoe1999AlzheimersDisorder, Mathotaarachchi2017IdentifyingImaging} AD takes a significant toll on patients'
daily lives, causing a progressive decline in their cognitive abilities,
including memory, language, behavior, and problem solving. \cite{Reisberg1985ClinicalDisease, Humpel2011IdentifyingDisease, Liu2013ApolipoproteinTherapy, Mu2011AdultDisease} Changes to AD patients' cognitive abilities
often start slowly and become more rapid over time. \cite{Teri1995CognitiveDecline, Moore2018RandomData}
Doctors and other caregivers monitor the progression of AD in patients
by evaluating the degree of decline in the patients' cognitive abilities
\cite{Neugroschl2011AlzheimersSeverity.}, which are often divided into 3 general categories: cognitively
normal (NL), mild cognitive impairment (MCI), and dementia.
\cite{Liu2013ApolipoproteinTherapy, Marinescu2018TADPOLEDisease} Patients with MCI and dementia both suffer from reduced
cognitive abilities, but MCI has a less severe effect on everyday
activities, and patients suffering from dementia often have additional
symptoms such as trouble with reasoning or impaired judgment.
\cite{Singanamalli2017CascadedFeatures, Gamberger2017IdentificationDisease}

Unfortunately, there is no cure for AD at this time, and progress on
identifying a cure has been slow. \cite{Marinescu2018TADPOLEDisease} None of the five medications
currently approved by the FDA to treat AD have been shown to delay or
halt its progression. \cite{Neugroschl2011AlzheimersSeverity.} Instead, they only temporarily improve
patients' symptoms. \cite{Neugroschl2011AlzheimersSeverity.,2016TheDisease, AlzheimersAssociation2010WhatDisease} And, the most recently
approved medication is just the combination of two existing drugs for
treating AD, donepezil and memantine, which were approved by the FDA 23
and 16 years ago, respectively. \cite{AlzheimersAssociation2010WhatDisease} Despite considerable efforts to
find a cure for AD, there is a 99.6\% failure rate of clinical trials
for AD drugs. \cite{2016TheDisease, Cummings2014AlzheimersFailures} In early 2018 alone, two groups ended
their AD clinical trials because their drugs failed to prevent the
progression of AD. \cite{Mumal2018PoorEarly, Cortez2018MerckFails} The difficulty in finding treatments
for AD is most likely a combination of uncertainty over the cause of AD
and the fact that AD patients cannot easily be identified at early
stages. \cite{Marinescu2018TADPOLEDisease, 2016TheDisease, 2018U.S.Administration} Even the FDA recognizes the importance
of identifying patients who are at risk of developing AD but who don't
have any noticeable cognitive impairment. \cite{Gottlieb2018StatementPrograms} For this reason, AD
research would benefit from the ability to use current medical data to predict the mental state of
patients in future years in order to identify patients who are good
candidates for clinical trials before they become symptomatic.
\cite{Mathotaarachchi2017IdentifyingImaging, Marinescu2018TADPOLEDisease, 2016TheDisease, Teipel2015MultimodalDetection, Oxtoby2017ImagingDisease}

Several different types of data have been identified that are
relevant to assessing the mental state of AD patients and the
progression of AD in general. One of the largest genetic risk factors
for AD is the presence of 1 or 2 copies of the $\epsilon$4 allele of the APOE
gene, which encodes a particular variant of the enzyme Apolipoprotein E.
\cite{Liu2013ApolipoproteinTherapy} Physical changes to the brain have also been shown to be
correlated with the progression of AD. For example, a decline in
neurogenesis in the hippocampus is one of the earliest changes to brain
physiology seen in AD patients and is thought to underlie cognitive
impairments associated with AD. \cite{Mu2011AdultDisease} The progression of AD also
accelerates the normal atrophy of brain tissue caused by aging, as
evidenced by increased enlargement of the ventricles of the brain over
time. \cite{Nestor2008VentricularDatabase} One study demonstrated a 4-fold difference in the rate of
ventricle enlargement in AD patients and normal controls over a
six-month interval. \cite{Nestor2008VentricularDatabase} Cognitive tests have also been widely used
for early detection of AD. \cite{Moore2018RandomData} Several commonly used tests, such as
ADAS11 and ADAS13, are based on the Alzheimer's Disease Assessment Scale
(ADAS), which is a brief cognitive test battery that assesses learning
and memory, language production, language comprehension, constructional
praxis, ideational praxis, and orientation. \cite{ADNIStaff2010ADNIManual, Kueper2018TheReview.} ADAS11
scores range from 0 to 70, and ADAS13 scores range from 0 to 85, with
higher scores indicating more advanced stages of AD. \cite{Kueper2018TheReview.} Similar
cognitive tests, such as the Mini-Mental State Examination (MMSE), the
Rey Auditory Verbal Learning Test (RAVLT), and the Functional Activities
Questionnaire (FAQ) have also been used to assess the progression of AD
in individual patients. \cite{Wan2014IdentifyingAccess, Tekin2001ActivitiesInfluences} ADAS has been found to be more
precise than the MMSE \cite{Balsis2015HowCorrespond}, and the RAVLT only addresses verbal
recall \cite{Boone2005ComparisonPerformance}, thus providing less diagnostic information than either
of the other two. Similarly, the FAQ only assesses a patients' ability
to perform certain tasks \cite{Pfeffer1982MeasurementCommunity.} and therefore is more limited in scope
than the MMSE and ADAS.

In recent years, machine learning techniques have been applied to the
diagnosis of AD patients with great success. For example, Esmaeilzadeh
\emph{et al.} achieved an accuracy of 94.1\% using 3D convolutional
neural networks to diagnose AD on a dataset with 841 patients. \cite{Esmaeilzadeh2018End-To-EndIdentification}
Similar results were obtained by Long \emph{et al}., who used a support
vector machine to diagnose AD based on an MRI scan dataset (n = 427
patients; mean best accuracy = 96.5\%) \cite{Long2017PredictionDeformation} and Zhang \emph{et al}.,
who used MRI scans, FDG-PET scans, and CSF biomarkers to diagnose AD (n
= 202 patients; AD vs. NL accuracy = 93.2\%, MCI vs. NL accuracy =
76.4\%). \cite{Zhang2011MultimodalImpairment.} However, the focus of these earlier studies was to use
current medical data to diagnose a patient's present cognitive state, in
effect demonstrating that a computer can replicate a doctor's clinical
decision-making. What is needed is a way to use machine learning to
predict future diagnoses of AD patients. \cite{Marinescu2018TADPOLEDisease}

\section{Methods}
\subsection{ADNI Data}
Data used in the preparation of this article were obtained from the
Alzheimer's Disease Neuroimaging Initiative (ADNI) database
(adni.loni.usc.edu). The ADNI was launched in 2003 as a public-private
partnership, led by Principal Investigator Michael W. Weiner, MD. The
primary goal of ADNI has been to test whether serial magnetic resonance
imaging (MRI), positron emission tomography (PET), other biological
markers, and clinical and neuropsychological assessment can be combined
to measure the progression of mild cognitive impairment (MCI) and early
Alzheimer's disease (AD).

The ADNI study has been divided into several phases, including ADNI-1,
ADNI-GO, and ADNI-2, which started in 2004, 2009, and 2011,
respectively. ADNI-1 studied 800 patients, and each subsequent phase
included a mixture of new patients and patients from the prior phase who
elected to continue to participate in the study. The ADNI patient data
was pre-processed to flag missing entries and to convert nonnumeric
categories (such as race) into numeric data. Data was sorted into three
datasets (LB1, LB2, and LB4) based on criteria established by The
Alzheimer's Disease Prediction Of Longitudinal Evolution (TADPOLE)
Challenge (https://tadpole.grand-challenge.org/). \cite{Marinescu2018TADPOLEDisease} The LB2 and
LB4 datasets consist of data from 110 patients who participated in
ADNI-1, continued to participate in ADNI-GO/ADNI-2, and who were not
diagnosed with AD as of the last ADNI-1 time point. Specifically, LB2
contains all observations of these patients from ADNI-1, and LB4
contains all observations of these patients from ADNI-GO/ADNI-2. The LB1
dataset consists of ADNI data for all remaining patients (n = 1737).
Generally speaking, LB1 was used as a training and validation dataset,
while LB2 and LB4 were later used to test the ability of machine
learning models to predict the progression of AD on an independent
patient population.

\subsection{All-Pairs Technique}
Data from LB1 was then further processed using a novel methodology
developed for this project called the All-Pairs technique, which can be
summarized as follows: Let \(R\) be the number of patients in the
dataset and \(B\) be the number of biomarkers (or other clinical data) being evaluated as
features. For each patient \(P_{i}\) (\(1 \leq i \leq R\)), the ADNI
database includes \(L_{i}\) separate examinations by a physician. Then,
\(E_{i,j}\), the \(j\)th examination of the \(i\)th patient, can be
defined as a multidimensional vector as follows:

\[E_{i,j} = \lbrack d_{i,j},b_{i,j,1},b_{i,j,2},...,b_{i,j,B},c_{i,j}\rbrack\]

where \(d_{i,j}\) is the date of the examination, \(b_{i,j,k}\)
(\(1 \leq k \leq B\)) are different biomarkers (or other clinical data), and \(c_{i,j}\) is the
clinical state of the patient (normal, MCI, or dementia) as measured
during that examination. The All-Pairs technique transforms this
examination data to generate a feature array \(X\) and target array
\(Y\) that are used to train the machine learning models. Specifically,
for every \(i,j_{a},j_{b}\), where \(1 \leq i \leq R\) and
\(1 \leq j_{a} < j_{b} \leq L_{i}\), a row of \(X\) and a corresponding
cell of \(Y\) are calculated:

\begin{align*}
X_{\text{row}} & = \lbrack d_{i,j_{b}} - d_{i,j_{a}},b_{i,j_{a},1},b_{i,j_{a},2},\ldots,b_{i,j_{a},B},c_{i,j_{a}}\rbrack \\
Y_{\text{cell}} & = c_{i,j_{b}}
\end{align*}

This approach can be extended from pairs of examinations
(\(j_{a},j_{b}\)) to triplets (\(j_{a},j_{b},j_{c}\)) as follows:

\begin{align*}
\begin{split}
X_{\text{row}} & = \lbrack d_{i,j_{c}} - d_{i,j_{b}},d_{i,j_{b}} - d_{i,j_{a}},b_{i,j_{b},1},\ldots,\\&\qquad b_{i,j_{b},B},c_{i,j_{b}},b_{i,j_{a},1},\ldots,b_{i,j_{a},B},c_{i,j_{a}}\rbrack
\end{split}
\\[2ex]
Y_{\text{cell}} & = c_{i,j_{c}}
\end{align*}

Individual rows and cells are then assembled to create the \(X\) and
\(Y\) arrays that are used for training of the machine learning model as
well as cross-validation studies.

\begin{table*}
    \centering
    \begin{tabularx}{\textwidth}{llX}
\toprule
\textbf{Category} & \textbf{Feature Name} &
\textbf{Description}\tabularnewline
\midrule
\rowcolor{lightgray}
Genetic & APOE4 & The APOE4 biomarker measures the number of copies (0,
1, or 2) of the $\epsilon$4 allele of the gene encoding the enzyme Apolipoprotein
E. This allele is the strongest genetic risk factor for AD.
\cite{Liu2013ApolipoproteinTherapy}\\
& Hippocampus & \\

\multirow{-2}{*}{Physical} & Ventricles/ICV & \multirow{-2}{=}{Physical measurements, like the volume of the
hippocampus \cite{Mu2011AdultDisease} and the ventricles \cite{Nestor2008VentricularDatabase}, can be used to assess
the effect AD has on the brain.}\\

\rowcolor{lightgray} & ADAS11 & \\

\rowcolor{lightgray} & ADAS13 & \\

\rowcolor{lightgray} & FAQ & \\

\rowcolor{lightgray} & MMSE & \\
\rowcolor{lightgray}
\multirow{-5}{*}{Behavioral} &  RAVLT (4 types) & \multirow{-5}{=}{Cognitive tests have been widely applied for early
detection of AD. \cite{Gainotti2013NeuropsychologicalDisease}}\\

 & Race & \\

\multirow{-2}{*}{Demographic}& Age & \multirow{-2}{=}{Demographic data can be used to account for general trends in a population.}\\
\bottomrule
\end{tabularx}
\caption{Features from the ADNI Dataset Used in this Study}
\label{tab:featUsed}
\end{table*}

\subsection{Evaluation using LB2 and LB4 Datasets}

In the case of LB2, each examination can be characterized by the same
vector \(E_{i,j}\) as used for the LB1 dataset. However, because the LB2
data is only used to test the machine learning models and not for
training, no comparison between examinations is performed. Instead,
\(E_{i,j}\) is transformed into the input vector by replacing the
\(d_{i,j}\) term with a time variable that represents the number of
months into the future that the machine learning model should make a
prediction. Specifically, for each patient \(P_{i}\) in LB2, the machine
learning algorithm is applied to a series of input vectors of the form:

\[\lbrack t,b_{i,j,1},b_{i,j,2},\ldots,b_{i,j,B},c_{i,j}\rbrack\]

where \(t\) is the time variable. Input vectors are generated based on
each patient's last three examinations in LB2, or for all of the
examinations if the patient has less than three. The probabilities
calculated by the machine learning algorithm based on these examinations
are averaged to generate predicted probabilities for patient \(P_{i}\)
at time\(\text{\ t}\). When comparing the model's predictions against
the actual diagnoses in LB4, \(t\) is set to the time difference between
the applicable exam in LB2 and the applicable exam in LB4. For the time
courses shown in Figures \ref{fig:predict} and \ref{fig:curves}, \(t\) is set to an integer between 1
and 84, inclusive.

\subsection{Features}
The features analyzed by the machine learning models consisted
of 13 biomarkers or other types of clinical data present in the ADNI dataset, all of which have been cited
in published papers as correlating with AD progression. These 13
features are summarized in Table \ref{tab:featUsed} and include genetic biomarkers
(APOE4), physical biomarkers (ventricular volume/ICV ratio,
hippocampal volume), the results of behavioral tests (ADAS13 and ADAS11
scores, FAQ score, MMSE score, and 4 types of RAVLT scores), and basic
demographic information (age and race). As described above, two
additional features that were generated during pre-processing of the
data using the All-Pairs technique were also included in the models,
namely, the clinical diagnosis at the earlier of two examinations and
the time difference between examinations.

\section{Results}
\subsection{Model Performance}
\begin{figure}[H]
    \centering
    \includegraphics[width=\columnwidth]{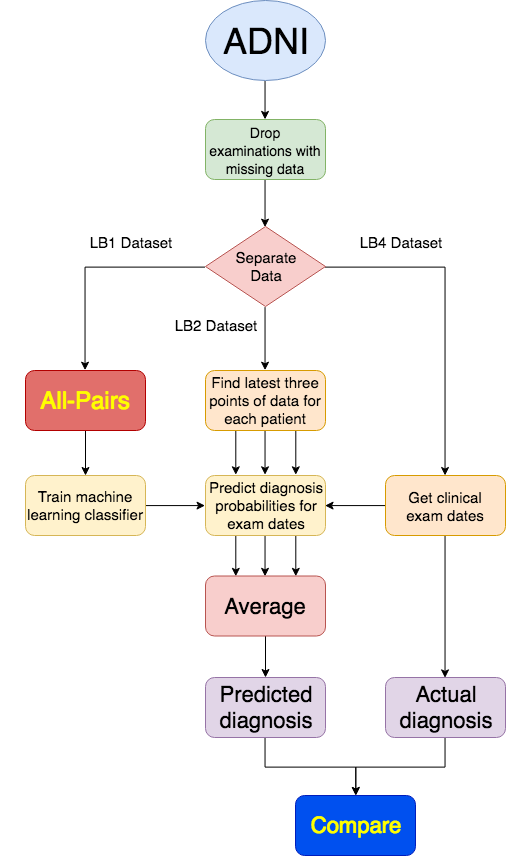}
    \caption{Project Flowchart}
    \label{fig:flowchart}
\end{figure}

A flowchart summarizing the overall methodology for training and
evaluating machine learning models is shown in Figure \ref{fig:flowchart}.

Various machine-learning classifiers, including support vector machines,
logistic regression, gradient boosting classifiers, random forests,
multilayer perceptron neural networks, and recurrent neural networks,
were implemented using the Python libraries \emph{Scikit-learn} and
\emph{Keras} (backed by \emph{TensorFlow}). Each classifier was then
evaluated on the processed data derived from LB1 using 7-fold
cross-validation.

\begin{table*}[!p]
    \centering
    \rowcolors{1}{}{lightgray}
    
    \begin{tabularx}{\textwidth}{lXccc}
    \toprule
    
\textbf{Variable} & \textbf{Meaning} & {\thead{8-feature\\training group}}
& {\thead{11-feature\\training group}} & {\thead{15-feature\\training
group}}\tabularnewline
\midrule

DX & Earlier Diagnosis & X & X & X\tabularnewline
ADAS13 & 13-item Alzheimer's Disease Assessment Scale & X & X &
X\tabularnewline
Ventricles & Ventricular Volume & X & X & X\tabularnewline
AGE & Age & X & X & X\tabularnewline
FAQ & Functional Activities Questionnaire & & X & X\tabularnewline
PTRACCAT & Race & X & X & X\tabularnewline
Hippocampus & Hippocampal Volume & X & X & X\tabularnewline
APOE4 & \# of APOE4 Alleles & X & X & X\tabularnewline
MMSE & Mini-mental state examination & & X & X\tabularnewline
ADAS11 & 11-item Alzheimer's Disease Assessment Scale & & X &
X\tabularnewline
RAVLT\_immediate & Rey Auditory Verbal Learning Test --- Total number of
words memorized over 5 trials & & & X\tabularnewline
RAVLT\_learning & Rey Auditory Verbal Learning Test --- Number of words
learned between trial 1 and trial 5 & & & X\tabularnewline
RAVLT\_forgetting & Rey Auditory Verbal Learning Test --- Number of
words forgotten between trial 5 and trial 6 & & & X\tabularnewline
RAVLT\_perc\_forgetting & Rey Auditory Verbal Learning Test ---
Percentage of words forgotten between trial 5 and trial 6 & & &
X\tabularnewline
N/A & Time Difference & X & X & X\tabularnewline
\bottomrule
\end{tabularx}
 \caption{Features from the ADNI Dataset Used to Train the MLP and RNN}
\label{tab:features}
\vspace*{\floatsep}

%\begin{figure*}
    \centering
    \includegraphics[width=0.8\textwidth]{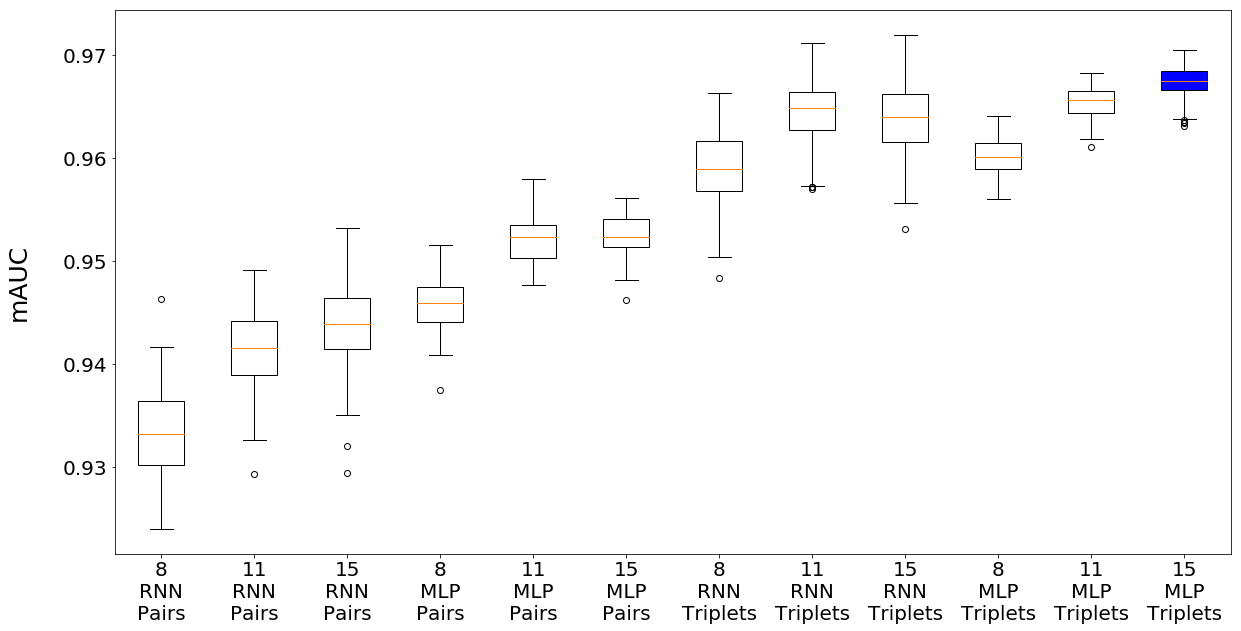}
    \captionof{figure}{Performance of Various Machine Learning Models (100 Random
Splits of 1737-Patient Training Dataset)}
    \label{fig:boxLB1}
%\end{figure*}%

\end{table*}

The effectiveness of each classifier was measured using a specialized
version of an ROC-AUC (Receiver Operating Characteristic Area Under the
Curve) score for multiclass classification (``mAUC score''), as
previously described by Hand and Till. \cite{Hand2001AProblems} The ROC-AUC score is a
balanced metric for classifiers that considers both the true positive
rate (percentage of actual positives that are called correctly) and the
false positive rate (percentage of actual negatives that are called
incorrectly). The mAUC variant of this score takes all ordered pairs of
categories \((i,\ j)\), measures the probability that a randomly
selected element from category \(i\) would have a higher estimated
chance of being classified as category \(i\) than a randomly selected
element from category \(j\), and averages all of these probabilities. A
classifier that works perfectly would have an mAUC score of 1; a
classifier that guessed randomly would result in an mAUC score of 0.5.

In the end, two classifiers, a multilayer perceptron implemented in
\emph{Scikit-learn} (``MLP'') and a recurrent neural network implemented
in \emph{Keras} (``RNN'') were found to have the best performance in the
cross-validation studies. Both of these classifiers are types of neural
networks. An MLP consists of a layer of input nodes, a layer of output
nodes, and one or more hidden layers between the input and output
layers. Each input vector is fed into the input nodes, and the value of
each node in every other layer is dependent on the values of the nodes
in the previous layer. Like an MLP, an RNN consists of multiple nodes
organized into layers, but the outputs of some of the hidden layers are
fed back into the same layer so that earlier input vectors can influence
the outputs for later input vectors. This allows the RNN to ``remember''
earlier inputs, which has been shown to be particularly useful when
analyzing data consisting of multiple observations taken at different
time points. \cite{Ghazi2018RobustModeling, Nguyen2018ModelingNetworks}

To further investigate the MLP and the RNN and see if the performance of
the MLP and RNN could be further improved by optimizing the training
protocol, six variants of each of the neural networks were generated to
examine the effects of changing the number of features being examined (8, 11, and 15) and whether the
All-Pairs technique was applied to pairs of patients' doctors' visits or
triplets of visits. Table \ref{tab:features} details which features were included in the
8, 11, and 15-feature training groups. The columns in Table \ref{tab:features} labelled
``Variable'' and ``Meaning'' provide the name of each feature as it
appears in the ADNI dataset as well as the corresponding description for
each feature, respectively.

Table \ref{tab:cv} shows the resulting mAUC scores after training and testing the
12 models on LB1-derived data using 7-fold cross-validation, as well as
the standard deviations for the testing mAUC scores. The best performing
model was the MLP trained on 8 features and triplets of time points, as
it had the highest testing mAUC score. This model is highlighted in blue
in Table \ref{tab:cv}. However, the differences among all of these models are
quite small and in many cases well within 1 standard deviation (s.d.),
suggesting that some of the variability between the scores might be due
to random chance rather than actual differences in predictive
performance. Additionally, because the training and testing of neural
networks, as well as the cross-validation process, relies to some extent
on algorithms that utilize random numbers, the scores in the table
will change slightly each time that the models are subjected to
cross-validation.%
\begin{table}[H]
    \centering
    \begin{tabularx}{\linewidth}{XS[table-format=1.4]S[table-format=1.4]S[table-format=1.4]}
    \toprule
    \textbf{Model} & {\thead{Train\\mAUC}} & {\thead{Test\\mAUC}} & {\thead{Test\\S.D.}}\\
    \midrule
    % 8 Features, RNN, pairs & 0.939015 & 0.931809\\
    % 11 Features, RNN, pairs & 0.948332 & 0.935697\\
    % 15 Features, RNN, pairs & 0.950447 & 0.931987\\ [1ex]
    % 8 Features, MLP, pairs & 0.948956 & 0.941818\\
    % 11 Features, MLP, pairs & 0.954321 & 0.952339\\
    % 15 Features, MLP, pairs & 0.954957 & 0.951480\\ [1ex]
    % 8 Features, RNN, triplets & 0.964932 & 0.915778\\
    % 11 Features, RNN, triplets & 0.972427 & 0.923033\\
    % 15 Features, RNN, triplets & 0.969669 & 0.935656\\ [1ex]
    % \rowcolor{blue!20} 8 Features, MLP, triplets & 0.960821 & 0.963074\\
    % 11 Features, MLP, triplets & 0.966245 & 0.950462\\
    % 15 Features, MLP, triplets & 0.966912 & 0.948111\\
    8 Features, RNN, pairs & 0.9390153857024401 & 0.9318087796495474 & 0.014449604988646525 \\
    11 Features, RNN, pairs & 0.9483324798652831 & 0.9356973815368063 & 0.012181959602566798 \\
    15 Features, RNN, pairs & 0.9504467369795584 & 0.9319867257788201 & 0.01482817212885263 \\ [1ex]
    8 Features, MLP, pairs & 0.9489560564623032 & 0.9418176340972437 & 0.007267225109912573 \\
    11 Features, MLP, pairs & 0.9543207931782944 & 0.9523385136108332 & 0.008269925390190534 \\
    15 Features, MLP, pairs & 0.9549574259667352 & 0.9514799501205221 & 0.008254362239343681 \\ [1ex]
    8 Features, RNN, triplets & 0.9649315752832309 & 0.9157779566811259 & 0.011844147702153517 \\
    11 Features, RNN, triplets & 0.9724271632434679 & 0.9230334145023289 & 0.011844147702153517 \\
    15 Features, RNN, triplets & 0.9696688415513751 & 0.9356560109163501 & 0.02358836540619911 \\ [1ex]
    \rowcolor{blue!20} 8 Features, MLP, triplets & 0.9608210978797398 & 0.9630736938756891 & 0.013335505798685537 \\
    11 Features, MLP, triplets & 0.9662454169851216 & 0.9504624280448501 & 0.012890860115087114 \\
    15 Features, MLP, triplets & 0.9669117573596644 & 0.9481111011887923 & 0.017367333009670582 \\
    \bottomrule
    \end{tabularx}
    \captionsetup{skip=10pt}
     \caption{Accuracy Metrics from Various Machine Learning Models (7-Fold Cross-Validation on 1737-Patient Training Dataset). MLP = multilayer perceptron implemented in \emph{Scikit-learn}; RNN = recurrent neural network implemented in \emph{Keras/Tensorflow}}
    \label{tab:cv}
\end{table}%

In order to assess the performance of these models in a more rigorous
manner, each of the 12 models was also evaluated on a series of random
splits of the preprocessed LB1 dataset. For each model, the preprocessed
LB1 dataset was randomly separated into a training dataset and testing
dataset using a 70:30 ratio, and this was repeated to create 100 pairs
of training and testing datasets that were then used to train and test
the model. The use of a large number of randomly generated splits
produces a distribution of mAUC scores that better reflects the overall
performance of the model and minimizes the effects of outliers. The
results are shown as box-and-whisker plots in Figure \ref{fig:boxLB1}. Using this more
rigorous approach, the model with the highest average mAUC score was an
MLP trained on data with 15 features and with triplets of patients'
examinations, which had an average mAUC score of 0.967 and a standard
deviation of 0.0016. This model is highlighted in blue in Figure \ref{fig:boxLB1}.%

\subsection{Prediction of Alzheimer's Disease Progression}
In order to assess real-world performance, select models were also
trained on the entire LB1 dataset (after processing with the All-Pairs
technique) and then evaluated on data derived from LB2 and compared to
actual diagnoses in LB4, asking whether early biomarkers and other early clinical data for the 110
patients in LB2 can predict their later diagnoses. The actual
examination dates in LB4 vary from patient to patient, but they
generally cover the 7-year period of ADNI-GO/ADNI-2. LB4 contains a
total of 417 examinations, or an average of 3.79 examinations per
patient. As in the cross-validation studies, the performance of each
model was assessed based on mAUC scores. Out of the 12 previously
discussed models, the 4 models that used 15 features were selected for
testing against LB2/LB4, namely:

1. MLP trained using pairs of examinations,

2. MLP trained using triplets of examinations,

3. RNN trained using pairs of examinations, and

4. RNN trained using triplets of examinations.

\begin{figure}
    \centering
    \includegraphics[width=\columnwidth]{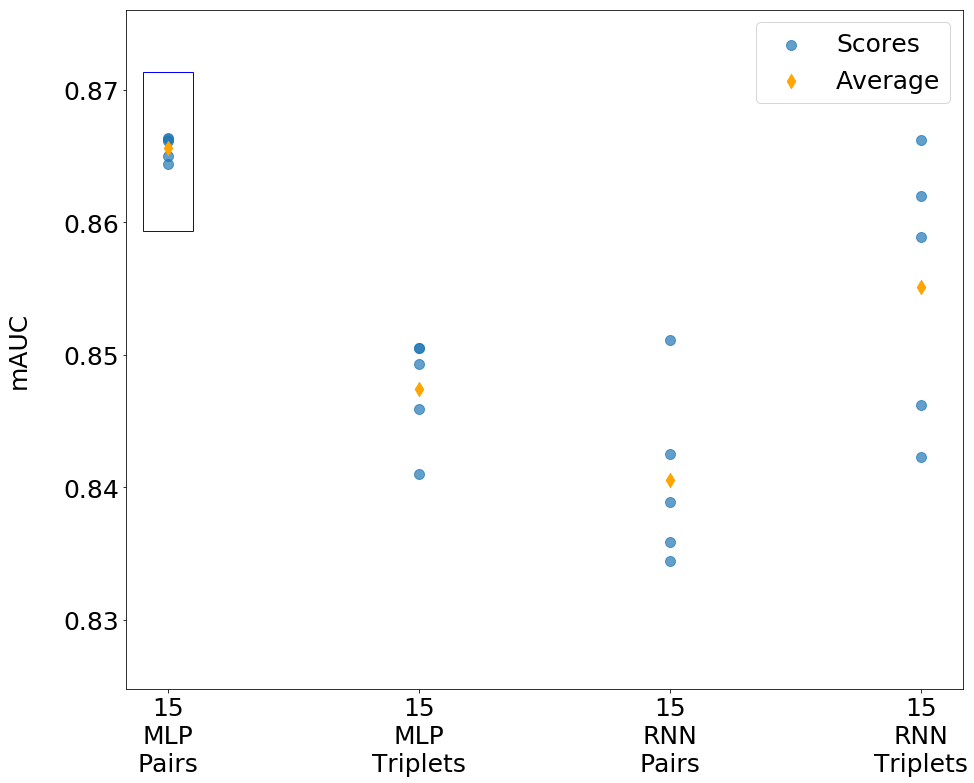}
    \caption{Performance of Various Machine Learning Models (110-Patient Testing Dataset, 5 Iterations)}
    \label{fig:dotsLB4}
\end{figure}

Hyper-parameter optimization was conducted for each of these models, and
each model was tested 5 times for each set of parameters to minimize the
impact of random variation inherent in the neural network training
process. In total, 27 different sets of parameters were tested for each
model, consisting of 3 possible values for the alpha parameter, 3
possible values for the learning rate, and 3 possible values for the
size of the hidden layers. Figure \ref{fig:dotsLB4} shows the results from the highest
performing version of each of the 4 models listed above.

The model and parameters with the best average mAUC score was the MLP
trained on 15 features using pairs of examinations, which achieved an
average score of 0.866 on the 110-patient test dataset. This model is
shown with a blue box in Figure \ref{fig:dotsLB4}. This result represents an improvement
over previously published work using the ADNI dataset. In particular,
Moore \emph{et al.} \cite{Moore2018RandomData} achieved an mAUC score of 0.82 with a random
forest classifier; Ghazi \emph{et al.} \cite{Ghazi2018RobustModeling} achieved an mAUC score
of 0.7596 with an RNN; and Nguyen \emph{et al.} \cite{Nguyen2018ModelingNetworks} achieved an
average mAUC score of 0.86 with an RNN together with forward-filling
data imputation.

\begin{figure}
    \centering
    \includegraphics[width=\columnwidth]{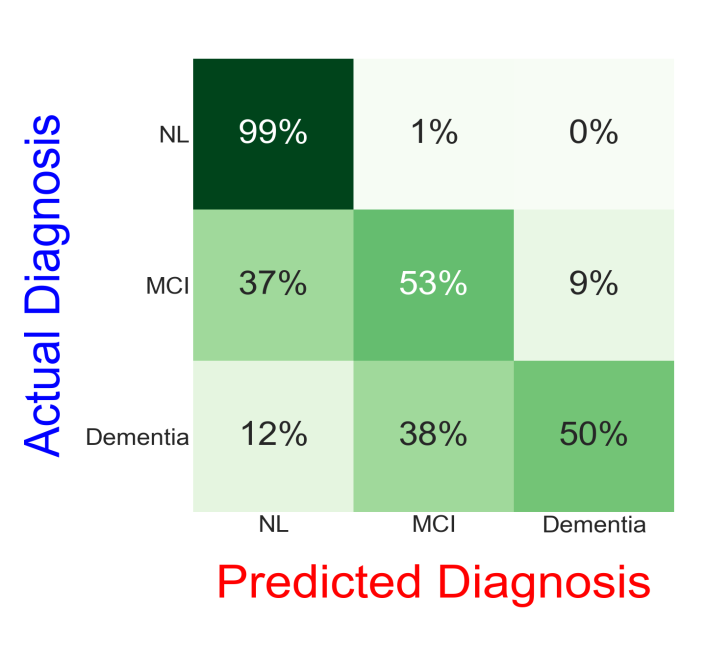}
    \caption{Confusion Matrix Comparing Predicted Diagnosis with Actual Diagnosis}
    \label{fig:cm}
\end{figure}

Figure \ref{fig:cm} depicts a confusion matrix for this best performing model,
which provides a visual indication of how well the diagnoses predicted
by the model line up with the actual diagnoses. The confusion matrix
reveals two types of mistakes occasionally made by the machine learning
model: predicting a cognitively normal diagnosis when a patient is
actually diagnosed with MCI and predicting an MCI diagnosis when a
patient is actually diagnosed with dementia. These mistakes may both
simply be the result of a small error in how the time variable is
applied by the algorithm, which creates a lag in the diagnosis
predictions.

\begin{figure}[!h]
    \centering
    \includegraphics[width=\columnwidth]{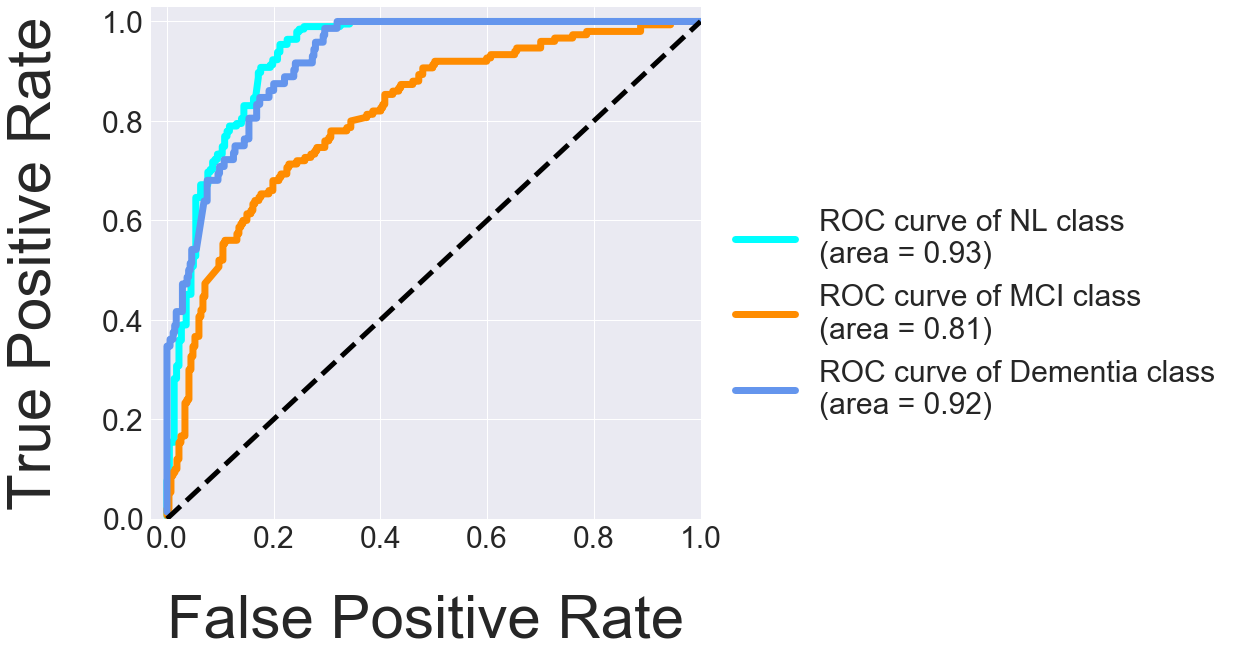}
    \caption{ROC Curves Based on Output of Neural Network}
    \label{fig:roc}
\end{figure}

Figure \ref{fig:roc} shows a group of Receiver Operating Characteristic (ROC) curves
based on the output of the best performing model, which measure how well
the model can separate two groups: patients with a particular diagnosis
and patients with one of the other diagnoses. Interestingly, the ROC
score for the MCI class is lower than the other two ROC scores,
suggesting that the model is having more difficulty separating patients
with MCI from the other two groups. Based on these individual ROC
scores, the model's average mAUC score (0.866) could be dramatically
increased if the model's ability to separate MCI patients from non-MCI
patients was improved.

\begin{figure*}
    \centering
    \includegraphics[height=9in]{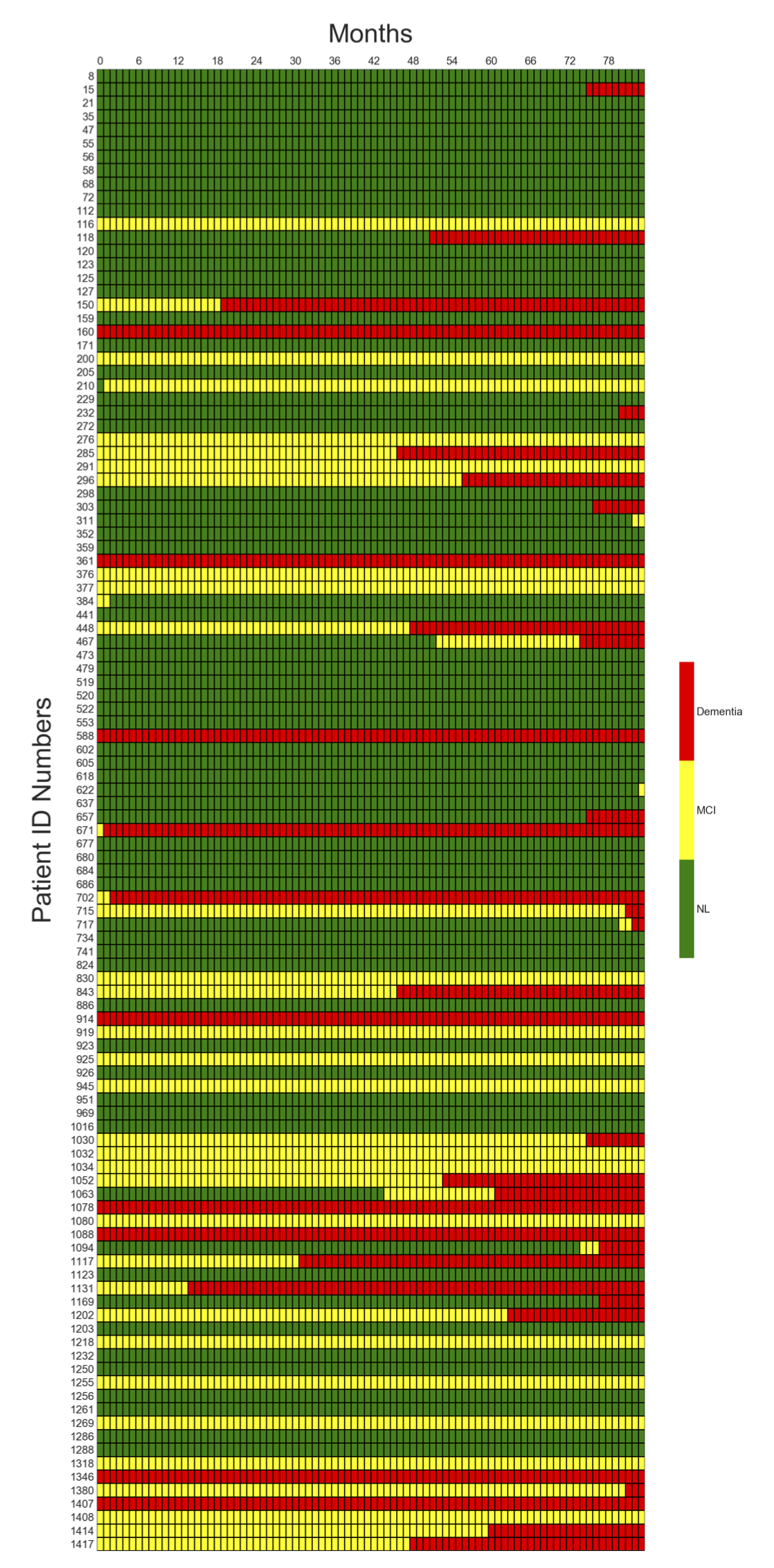}
    \caption{Month-by-Month Predicted Diagnosis for 110 Patients
over 7 Years}
    \label{fig:predict}
\end{figure*}

The best performing model was also used to predict the future diagnosis
of normal, MCI, or dementia for all 110 patients in the LB2/LB4 datasets
on a month-to-month basis over an 84-month (7-year) period, as shown in
Figure \ref{fig:predict}. The model predicted that some patients would remain normal
over the entire 7-year period, while others would progress from normal
to MCI to dementia.

\begin{figure*}[!t]
    \centering
    \includegraphics[width=\textwidth]{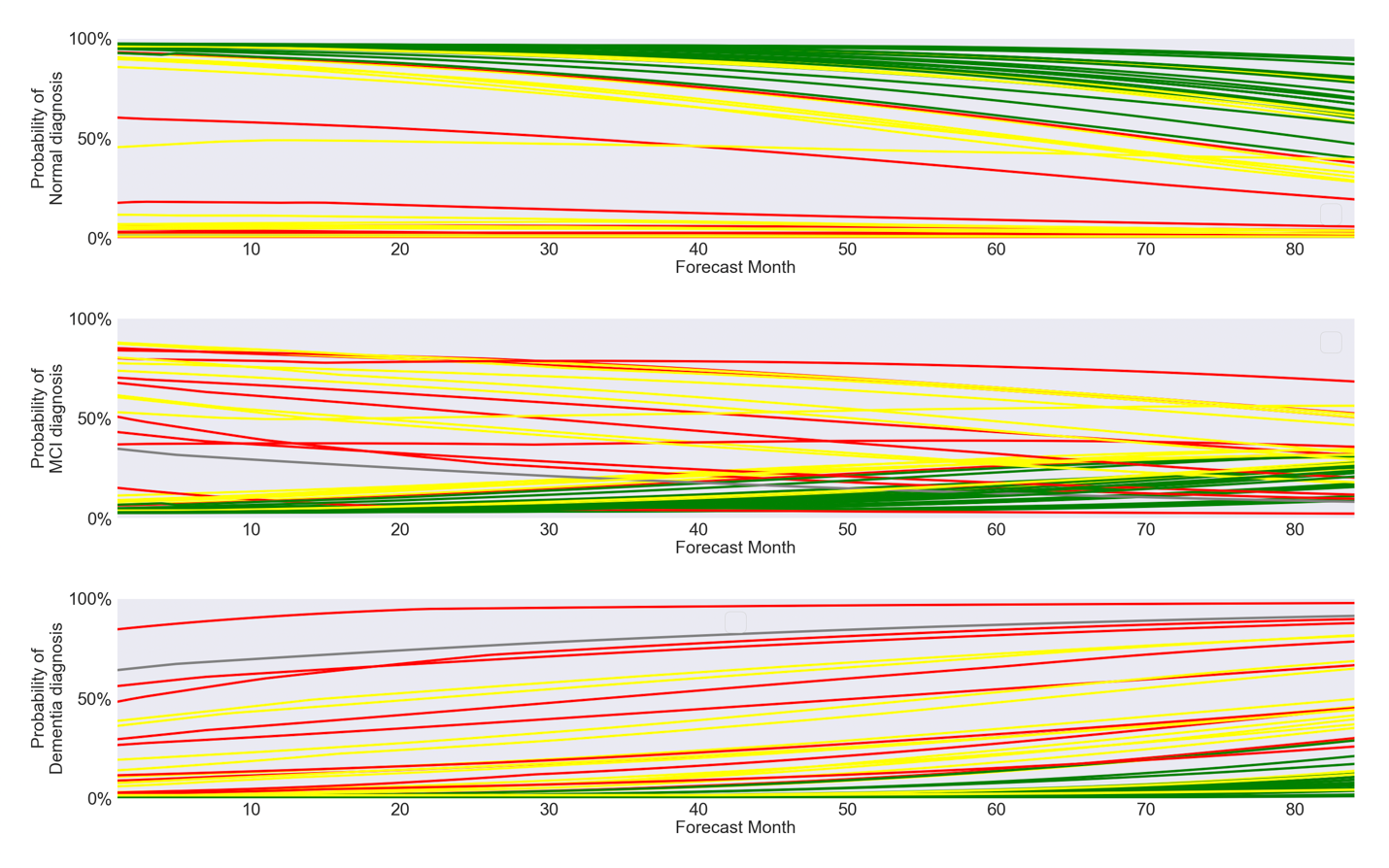}
    \caption{Predicted Likelihood of Diagnoses for 50 Random Patients over
7 Years}
    \label{fig:curves}
\end{figure*}

Time courses were also generated for a random subset of these patients
(n=50) showing how the likelihood of normal, MCI, or dementia diagnoses
is forecast to vary over the 84 months, as shown in Figure \ref{fig:curves}. The color
of each curve (green, yellow, or red) indicates the most severe
diagnosis actually received by the patient (normal, MCI, or dementia,
respectively) during the 7-year period. As can be seen in Figure \ref{fig:curves},
patients who remained normal over the entire 84-month period generally
received very low predicted probabilities of MCI or dementia diagnoses
from the model. Similarly, patients who were diagnosed with dementia at
some point during this period generally received high predicted
probabilities of dementia from the model.

\section{Conclusion}
Previous work published by others has shown that machine learning
algorithms can accurately classify a patient's current cognitive state
(normal, MCI, or dementia) using contemporaneous clinical data.
\cite{Esmaeilzadeh2018End-To-EndIdentification, Long2017PredictionDeformation, Zhang2011MultimodalImpairment.} This project has extended this previous work by
looking at how past and present clinical data can be used to predict a
patient's future cognitive state and by developing machine learning
models that can correlate clinical data obtained from patients at one
time point with the progression of AD in the future. Several of the
machine-learning models used in this project were effective at
predicting the progression of AD, both in cognitively normal patients
and patients suffering from MCI. Additionally, a novel All-Pairs
technique was developed to compare all possible pairs of temporal data
points for each patient to generate the training dataset. By comparing
data points at different points in time, the All-Pairs technique adds
time as a variable and therefore does not require fixed time intervals,
which are unlikely to occur in ``real-life'' data. \cite{Bernal-Rusiel2013StatisticalModels} These
techniques could be used to identify patients having high AD risk before
they are diagnosed with MCI or dementia and who would therefore make
good candidates for clinical trials for AD therapeutics. Since the
inability to identify AD patients at early stages is believed to be one
of the primary reasons for the frequent failure of AD clinical trials,
these techniques may help increase the chances of finding a treatment
for AD.

%%%%%%%%%%%% Supplementary Methods %%%%%%%%%%%%
%\footnotesize
%\section*{Methods}

%%%%%%%%%%%%% Acknowledgements %%%%%%%%%%%%%
\footnotesize
\section*{Acknowledgements}
Thank you to Jen Selby, Rachel Dragos, Ted Theodosopoulos, Daphne Koller, and Michael Weiner for reviewing this manuscript and providing me with valuable feedback, suggestions, and advice.

Data collection and sharing for this project was funded by the Alzheimer's Disease Neuroimaging Initiative (ADNI) (National Institutes of Health Grant U01 AG024904) and DOD ADNI (Department of Defense award number W81XWH-12-2-0012). ADNI is funded by the National Institute on Aging, the National Institute of Biomedical Imaging and Bioengineering, and through generous contributions from the following: AbbVie, Alzheimer’s Association; Alzheimer’s Drug Discovery Foundation; Araclon Biotech; BioClinica, Inc.; Biogen; Bristol-Myers Squibb Company; CereSpir, Inc.; Cogstate; Eisai Inc.; Elan Pharmaceuticals, Inc.; Eli Lilly and Company; EuroImmun; F. Hoffmann-La Roche Ltd and its affiliated company Genentech, Inc.; Fujirebio; GE Healthcare; IXICO Ltd.; Janssen Alzheimer Immunotherapy Research \& Development, LLC.; Johnson \& Johnson Pharmaceutical Research \& Development LLC.; Lumosity; Lundbeck; Merck \& Co., Inc.; Meso Scale Diagnostics, LLC.; NeuroRx Research; Neurotrack Technologies; Novartis Pharmaceuticals Corporation; Pfizer Inc.; Piramal Imaging; Servier; Takeda Pharmaceutical Company; and Transition Therapeutics. The Canadian Institutes of Health Research is providing funds to support ADNI clinical sites in Canada. Private sector contributions are facilitated by the Foundation for the National Institutes of Health (www.fnih.org). The grantee organization is the Northern California Institute for Research and Education, and the study is coordinated by the Alzheimer’s Therapeutic Research Institute at the University of Southern California. ADNI data are disseminated by the Laboratory for Neuro Imaging at the University of Southern California.

%%%%%%%%%%%%%%   Bibliography   %%%%%%%%%%%%%%
\normalsize
\printbibliography

%%%%%%%%%%%%  Supplementary Figures  %%%%%%%%%%%%
%\clearpage

%%%%%%%%%%%%%%%%   End   %%%%%%%%%%%%%%%%
%\end{multicols}  % Method B for two-column formatting (doesn't play well with line numbers), comment out if using method A
\end{document}